\pdfoutput=1
\documentclass[11pt]{article}

\usepackage{ACL2023}
\usepackage{times}
\usepackage{latexsym}

\usepackage[T1]{fontenc}
\usepackage[utf8]{inputenc}

\usepackage{multirow}
\usepackage{colortbl}
\usepackage{booktabs}
\usepackage{listings}
\usepackage{xcolor}
\usepackage{makecell}
\usepackage{microtype}
\usepackage{xurl}
\usepackage{tcolorbox}
\usepackage{amssymb}% http://ctan.org/pkg/amssymb
\usepackage{pifont}% http://ctan.org/pkg/pifont
\newcommand{\cmark}{\ding{51}}%
\newcommand{\xmark}{\ding{55}}%

\newcommand{\OURS}{FalseQA }
\newcommand{\OURSNOSPACE}{FalseQA}
\usepackage{inconsolata}
\usepackage{amsmath}
\usepackage{graphicx}
\usepackage{tablefootnote}
\usepackage{hyperref}

\newcommand{\ls}{\looseness=-1}
\newcommand{\std}[2]{$\text{#1}\pm\text{#2}$}  
\newcommand{\myemph}[1]{\cellcolor[HTML]{e1efd8}{#1}}
\definecolor{emphcolor}{rgb}{0.882, 0.937, 0.847}
\usepackage{float}
\title{Won't Get Fooled Again: Answering Questions with False Premises}

\author{Shengding Hu$^{1}$, Yifan Luo$^{2}$, Huadong Wang$^{1}\thanks{\quad Corresponding author: Huadong Wang (huadw2012@163.com)}$,\\
\textbf{Xingyi Cheng}$^{3}$\textbf{,  Zhiyuan Liu$^{1,4,5}$}\textbf{, Maosong Sun}$^{1,4,5}$\\
\textsuperscript{1}Dept. of Comp. Sci. \& Tech., Institute for AI, Tsinghua University, Beijing, China\\
Beijing National Research Center for Information Science and Technology\\
\textsuperscript{2} School of Cyberspace Security, BUPT
\textsuperscript{3}Tencent \\
\textsuperscript{4}Institute for Artificial Intelligence, Tsinghua University\\
\textsuperscript{5}International Innovation Center of Tsinghua University, Shanghai, China\\
{\tt hsd20@mails.tsinghua.edu.cn, yifanluo@bupt.edu.cn}
}

\begin{document}
\maketitle
\begin{abstract}
\ls Pre-trained language models (PLMs) have shown unprecedented potential in various fields, especially as the backbones for question-answering (QA) systems. However, they tend to be easily deceived by tricky questions such as ``How many eyes does the sun have?''. Such frailties of PLMs often allude to the lack of knowledge within them. In this paper, we find that the PLMs already possess the knowledge required to rebut such questions, and the key is how to activate the knowledge. To systematize this observation, we investigate the PLMs' responses to one kind of tricky questions, i.e., the false premises questions (FPQs).
We annotate a \OURS dataset containing 2365 human-written FPQs, with the corresponding explanations for the false premises and the revised true premise questions.
Using \OURSNOSPACE, we discover that PLMs are capable of discriminating FPQs by fine-tuning on moderate numbers (e.g., 256) of examples. PLMs also generate reasonable explanations for the false premise, which serve as rebuttals. Further replaying a few general questions during training allows PLMs to excel on FPQs and general questions simultaneously. Our work suggests that once the rebuttal ability is stimulated, knowledge inside the PLMs can be effectively utilized to handle FPQs, which incentivizes the research on PLM-based QA systems. The \OURS dataset and code are available at \url{https://github.com/thunlp/FalseQA}.

%Pre-trained language models (PLMs) have shown unprecedented potential in various fields, especially as the backbones for question-answering (QA) systems. However, they tend to be easily deceived by tricky questions such as ``How many eyes does the sun have?''. Such frailties of PLMs often allude to the lack of knowledge within them. In this paper, we find that the PLMs already possess the knowledge required to rebut such questions, and the key is how to activate the knowledge. To systematize this observation, we investigate the PLMs' responses to one kind of tricky questions, i.e., the false premises questions (FPQs). We annotate a FalseQA dataset containing 2365 human-written FPQs, with the corresponding explanations for the false premises and the revised true premise questions. Using FalseQA, we discover that PLMs are capable of discriminating FPQs by fine-tuning on moderate numbers (e.g., 256) of examples. PLMs also generate reasonable explanations for the false premise, which serve as rebuttals. Further replaying a few general questions during training allows PLMs to excel on FPQs and general questions simultaneously. Our work suggests that once the rebuttal ability is stimulated, knowledge inside the PLMs can be effectively utilized to handle FPQs, which incentivizes the research on PLM-based QA systems. The code and dataset will be publicly available.
\end{abstract}

\section{Introduction}

\ls Recent advances in pre-trained language models (PLMs) ~\cite{lewis-etal-2020-bart, raffel2020exploring, brown2020language, roller-etal-2021-recipes, han2021pre}  have achieved significant performance gains for various types of tasks, even surpassing human levels on language ability benchmarks~\cite{wang-etal-2018-glue, wang2019superglue,srivastava2022beyond}. The unprecedented ability of PLMs lays the foundation for various practical applications. For example, PLMs that exhibit general world knowledge and commonsense knowledge have the potential to serve as backbones for general-purpose question-answering models~\cite{tafjord2021general,guu2020retrieval}.
\begin{figure}[!tpb]
    \centering
    \includegraphics[width=0.9\linewidth]{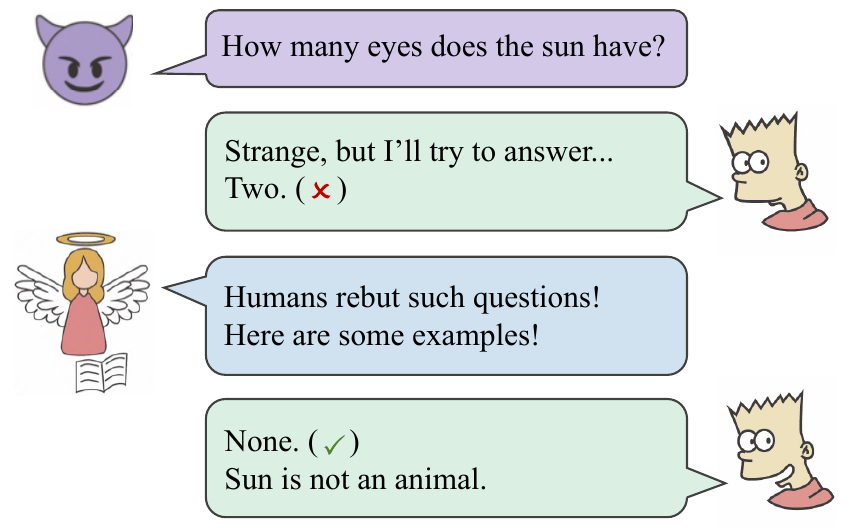}
    \caption{The rebuttal ability of PLMs can be activated by human rebuttal examples. }
    \label{fig:intrograph}
\end{figure}

\begin{table}[]
\vspace{-0.5em}
    \centering
\resizebox{\linewidth}{!}{
    \begin{tabular}{m{5.8cm}|m{3.3cm}}
    \toprule
      Question  & Answer\\
     \hline
     Sally’s favorite cow died yesterday. When will the cow be alive again? \tablefootnote{\href{https://blog.allenai.org/general-purpose-question-answering-with-macaw-84cd7e3af0f7}{AllenAI's Blog.}}   & in a few days. \\
    \midrule
     How many eyes does the sun have? \tablefootnote{\href{https://lacker.io/ai/2020/07/06/giving-gpt-3-a-turing-test.html}{Blog \textit{Giving GPT-3 a Turing Test}. }}&
     The sun has one eye. \\
     \bottomrule
    \end{tabular}
}
    \caption{Some previous examples that report the vulnerability of PLMs to tricky questions. More examples provided by this work are in Table~\ref{tab:direct_answer_examples_main}. }
    \vspace{-1.5em}
    \label{tab:intro_example}
\end{table}

\ls However, these PLM-based question-answering models have an intriguing paradox. On the one hand, they achieve high performance on normal questions raised by humans. For example, \textsc{UnifiedQA}~\cite{khashabi-etal-2020-unifiedqa} achieves state-of-the-art performance on many question-answering tasks. \textsc{Macaw}~\cite{tafjord2021general} can perform multi-angle question-answering and answer 75\% of the question in the Challenge300 dataset~\cite{tafjord2021general} correctly. On the other hand, they are vulnerable to tricky questions (see Table~\ref{tab:intro_example}). For example, \textsc{Macaw} answers one out of nine tricky questions correctly, while other models including GPT-3~\cite{brown2020language} fail all of them~\cite{tafjord2021general}. InstructGPT~\cite{ouyang2022training} also reports that it fails to identify instructions with false premises. These questions are easy to rebut for humans but pose an undeniable obstacle for PLMs\footnote{Although most PLMs fail, we found ChatGPT~\cite{chatgpt} satisfactorily answers these questions. Their training data is manually written by annotators and continuously updated using user queries, which might contain such questions. However, their data is not public. Our work provides the same possibility for general PLMs, even the much smaller ones.}. The inability to rebut also results in the misalignment~\cite{kenton2021alignment} of language models to human expectations.

Without careful investigation, this paradox could easily lead to the conclusion that PLMs lack the world or commonsense knowledge to rebut these questions. Although it's crucial for the PLMs to embed as much general knowledge as possible, we provide a pilot experiment to find out that the PLMs already possess the knowledge required for the tricky questions which they fail (see Section~\ref{sec:pilot}). As a consequence, we hypothesize that the knowledge in current PLMs is \textit{enough} for handling a large portion of tricky questions. However, this knowledge is \textit{not activated}.

To support our hypothesis, we take a close look at these tricky questions. Most of these tricky questions contain false premises. For example, in the question ``\emph{How many eyes does the sun have?}'', the questioner must presume that ``\emph{the sun can have eyes}'' in order to make the query about the quantity meaningful. These questions are called False Premise Questions (FPQs). Such false premises always violate human knowledge or logic and rarely appear in the natural text, thus leading to an out-of-distribution generalization gap for the PLMs.

\ls Targeting to fill the gap between the natural text and FPQs, we present the first specialized dataset of FPQs, dubbed as \OURS dataset. Specifically, we first systematically categorize the false premises to ensure the coverage of the dataset. Then we ask human annotators to manually compose the FPQs, as well as explanations for the false premises. The annotators are also asked to edit the false premise questions into true premise questions (TPQs) using minimal modification, with which the PLMs are less prone to learn shortcuts from the format of FPQs.  

% and demonstrate PLMs' potential at these questions, 
Based on \OURS dataset, we first conduct systematic experiments on the PLMs’ discrimination and rebuttal ability of FPQs.
We reach three essential conclusions: ~(1) PLMs of different types and scales can distinguish the FPQs from TPQs, and scaling effect~\cite{kaplan2020scaling} also holds for \OURSNOSPACE. (2) PLMs can give reasonable explanations for the false premises, which can serve as rebuttals. (3) The number of FPQ examples needed to activate the PLM's rebuttal ability is moderate. For example, 256 FPQs can result in more than 70\% accuracy for models larger than 1B. And for some larger PLMs, in-context learning with a few examples can also activate the ability. Then we consider the practical scenario where the models need to handle both FPQs and general questions. We demonstrate that a simple but effective  data replay method can help
% However, since the model is adapted to \OURSNOSPACE's 
% adversarial question distributions, performance degradation on normal questions like ARC-DA~\cite{bhakthavatsalam2021think} are observed. Aiming at
mitigate the catastrophic forgetting of general questions, where the model discriminates 86.7\% FPQs in \OURS and only rebuts 1.4\% general questions. These results lead to optimism that PLMs can be used as the backbones of a practical question-answering system that is robust to tricky questions.

\begin{table*}[!htbp]
\vspace{0cm}
\centering
\resizebox{\textwidth}{!}{
    \begin{tabular}{m{2.2cm}|c|p{9.3cm}|c|l}
\toprule 
  Model  & \# Para & \multicolumn{1}{>{\centering\arraybackslash}c|}{Reply} & Correctness & \multicolumn{1}{>{\centering\arraybackslash}c}{Ablation}\\
 \midrule
\multicolumn{4}{c}{\cellcolor[HTML]{EAEAEA} Are tigers' eggs bigger than chicken's eggs?} & \multicolumn{1}{>{\centering\arraybackslash}c}{\cellcolor[HTML]{EAEAEA} Do tigers lay eggs?} \\
\midrule
  Bloom & 176B &  No, they are not. The eggs of tigers are smaller than ... & \xmark & No tigers lay eggs.\\
  OPT & 175B & Yes. & \xmark & No, they don't. Tigers are mammals. \\
  Jurassic-1 & 178B  &  No, tigers' eggs are the same size as chicken's eggs. & \xmark & Tigers give birth to live young.\\
  GPT-3 & 175B & Yes, tiger eggs are larger than chicken eggs. Tiger eggs can ... & \xmark & No, tigers do not lay eggs ...\\
  \textsc{Macaw} & 11B & Yes. & \xmark & Yes.\\ % lyf edit
  \textsc{Macaw}-FPQ & 11B & Tigers don't lay eggs, chickens do. & \cmark & Tigers are carnivores and do not lay eggs. \\ % lyf edit
\midrule 
 \multicolumn{4}{c}{\cellcolor[HTML]{EAEAEA} Which Chinese classic do you like best, Faustus or Hamlet? } &\cellcolor[HTML]{EAEAEA} Are Faustus and Hamlet Chinese classics?\\
 \midrule 
  Bloom & 176B &   I like Faustus best. I like Hamlet, but I like Faustus best. & \xmark &  No, they are not. They are Western ...\\
  OPT & 175B & I like both. & \xmark & No. \\
  Jurassic-1 & 178B  & I like Faustus best. Faustus is a man, and Hamlet is a ghost. & \xmark & No, they are not.\\
  GPT-3 & 175B & I personally prefer Faustus, as it is a classic Chinese play ... & \xmark & No, Faustus and Hamlet are not ... \\
  \textsc{Macaw} & 11B & Faustus. & \xmark & No.\\ % lyf edit
  \textsc{Macaw}-FPQ & 11B & Faustus and Hamlet are European classics. & \cmark & Faustus and Hamlet are European classics.\\ % lyf edit
  \bottomrule
    \end{tabular}
}
 \caption{A case study of PLMs' responses to FPQs and the questions that ask the correctness of the premises. The questions are from the test split of \OURS dataset. See Appendix~\ref{app:dirrect_answer} for more examples.}
\label{tab:direct_answer_examples_main}
\vspace{-0.3cm}
\end{table*}

\section{Related Work}
Three groups of research are related to our work: direct question answering dataset,  question unanswerability, and question premise verification.

\textbf{Direct Question Answering Dataset.} For the most practical scenario of the question-answering system, the candidate answers are absent. Therefore, direct question answering (DQA), as a counterpart to extractive QA~\cite{rajpurkar-etal-2016-squad} or multiple-choice QA~\cite{lai-etal-2017-race} has received increased attention. Natural Questions~\cite{kwiatkowski2019natural} collects the queries sent to the google search engine. ARC-DA~\cite{bhakthavatsalam2021think} proposes modifying a reasoning-based multiple choice QA into DQA format. ~\citet{tafjord2021general} manually compose Challenge300 dataset which is still challenging to powerful models such as GPT-3 and \textsc{Macaw}. Our dataset can be seen as a direct question-answering dataset with explanations. However, the question distribution is radically different from the questions in natural corpora, serving as an adversarial scenario for DQA models.

\textbf{Question Unanswerability.} Tricky questions are unanswerable questions.  Previous works~\cite{raina2022answer, rajpurkar-etal-2018-know, asai-choi-2021-challenges, davis2020unanswerable} confirm the existence of unanswerable questions in existing benchmarks, including SQuAD~\cite{rajpurkar-etal-2016-squad}, Natural  Questions~\cite{kwiatkowski2019natural}, VQA~\cite{antol2015vqa}, etc. Most unanswerable questions in these benchmarks are due to missing information in the context provided to the questions. However, \OURS contains questions that are out of natural text distribution, and are unanswerable due to misleading false premises.
% Typical solutions in the previous work include training with the annotated answerable or unanswerable labels. We take similar perspective that once we provide the ``hint'' the model to rebuttal using a few examples, the model will handle tricky questions successfully. However, the unanswerability of intentional tricky questions are risen due to abnormal 

% can be categorised as lack of information~\cite {asai-choi-2021-challenges}, incorrect personal experience~\cite{yen2021unanswerable}. Generally, user prefers a refusal to answer the question to a false answer. Our work take similar perspective and encourage the model to give rebuttal to false premise questions. 

\ls\textbf{Question Premise Verification.} Answering FPQs has been studied before the deep learning era~\cite{kaplan1978indirect}. In recent PLM-based question-answering research, relevant efforts use external knowledge to verify the correctness of the question premise. For example, ~\citet{kim-etal-2021-linguist} studies the FPQs in Natural Questions~\cite{kwiatkowski2019natural}. A concurrent work~\cite{min2022crepe} further gathers the 8400 Reddit questions and annotated the false premises among 25\% of them. The correctness of the premises in their datasets requires expert knowledge or context to determine. Therefore, they use retrieval-augmented language models~\cite{krishna-etal-2021-hurdles} or external knowledge base to provide information for the premise classification, and both reach the conclusion that discovering and explaining those prepositions that require expert knowledge is challenging. However, it remains elusive whether PLMs without external assistance can discover and rebut the tricky questions that require only general knowledge and are straightforward for humans. We propose the first manually written dataset for FPQs and support our hypothesis through experiments that the inability of PLMs for FPQs can be mitigated when giving them examples.

% These two papers might support the inability of PLMs towards FPQs. However, more in-depth analysis on GPT-3 \footnote{\url{https://twitter.com/nicklovescode/status/1284050958977130497}} demonstrates that using an uncertainty prompt that hint the existence of FPQs can partly activate the identification ability of tricky questions without external assistance~\footnote{The textual instructions contains one or more tricky questions with the answer ``\emph{yo be real}'', then when encounter similar tricky questions, the answer will be ``\emph{yo be real}''.}. 

% Although the textual indications given to GPT-3 are promising, three gaps remain in the pursuit of human-like responses to FPQs. First, a comprehensive FPQ dataset is absent, as FPQs rarely appear in natural text corpora.
% Second, the existence of intrinsic discrimination ability of FPQs in a broader range of PLMs other than GPT-3 remains unknown. Third, it remains unknown whether large PLMs can generate human-like rebuttals and justify them, rather than just give predictions.

% \vspace{-3mm}
\section{Preliminaries}
% \vspace{-3mm}
In this section, we introduce the definition of FPQ and the pilot experiment on PLMs about FPQs.

\subsection{False Premise Questions}
When questioning, humans usually assume that some facts are shared and endorsed by the questioner and the answerer. Such facts  are the premises of the question. For example, in the question ``\textit{How many eyes does the sun have?}'', the target of the question is the number of eyes, which assumes the correctness of the fact ``\textit{The sun has eyes}''.

\ls In general, a fact can be expressed by relational triples, where each relational triple takes the form of \texttt{<subject, predicate, object>}. A question is asking for the missing part in one relational triple. For example, the above question can be expressed as nested triples as \texttt{<triple, quantity, ?>}, where \texttt{triple = <sun, has\_property, eye>}. We define the complete relational triple as the support triple. Then a false premise problem is one whose support triples are not correct. In the above example, \texttt{<sun, has\_property, eye>} is false under real-world background, thus any question that builds on this triple contains false premises. By this definition,  ``Does the sun have eyes?'' is not an FPQ, since it does not assume  \texttt{<sun, has\_property, eye>} to be true. In fact, PLMs know the authenticity of such triples well. However, they can't answer FPQs built upon these triples. 

% \begin{figure*}
\begin{table*}[]
\vspace{-0.5cm}
    \centering
\resizebox{0.98\textwidth}{!}{
    \begin{tabular}{m{2.5cm}|c|m{6.5cm}|m{9cm}}
\toprule
{\textbf{Category}} & \textbf{Fraction (\%)}& \multicolumn{1}{>{\centering\arraybackslash}c|}{\textbf{Description}} & \multicolumn{1}{>{\centering\arraybackslash}c}{\textbf{Example}}\\
\midrule
\multicolumn{4}{c}{\cellcolor[HTML]{D0D0D0}{\textbf{Error Types}}}  \\
\midrule
Property & 23.2 & The entity does not has the property. & How long has the Sun been transparent?\\
\midrule
Action & 19.7 & The entity can not perform the action. & How far can a fish walk on the street? \\
\midrule
Scope & 19.6  & A fact is not valid in the scope. & Who is the villain who fought Harry Potter in A Song of Ice and Fire? \\
\midrule
Entity & 11.3 & The entity can not exist. & What's the most common color of human's wings?\\
\midrule
Event  & 8.3 & The event didn't happen in the history. &  When did Zuckerberg start Google?\\
\midrule
 Logic & 6.7& Contain logically conflicting statements. & 
How to sit down while walking? \\
\midrule
 Causality  & 5.6 & Does not follow causality. & Why the more water you drink, the more thirsty you are?   \\
\midrule
 Index  & 4.6 & The specified index is out of an entity list. & What is the 50th largest province in China?\\
\midrule
\multicolumn{4}{c}{\cellcolor[HTML]{D0D0D0}{\textbf{Question Formats}}}  \\
\midrule
Descriptive &  29.6 & The question needs descriptive answer.  & Why carbon dioxide is composed of oxygen?  \\
\midrule
Factual & 28.1 &  The question seeks factual information. &  When did China become a member of the EU?\\
\midrule
Enumerative & 12.3 & The answer is a list of items. & List three vegetables that tigers feed on. \\
\midrule
Selective &  10.7 &  The answer candidates are provided. & Which one is the right behave in the theatre? Fight or disrupt the show? \\
\midrule
Hypothetical & 9.0 & The question contains a conditional clause. & When should I go if I want to see the fog on Jupiter? \\
\midrule
Affirmative & 8.5 & The question requires a yes-or-no answer. &  Do people eat diamond because it comes with mutiple nutrition? \\ 
\bottomrule
    \end{tabular}
}
\vspace{-0.5em}
    \caption{The categorization and examples of FPQ questions. We omit the ``Other'' category in this table.}
    \label{tab:error_types}
\vspace{-0.7em}
\end{table*}
% \end{minipage}
% \end{figure*}
% \vspace{2cm}

\subsection{PLMs' original responses to FPQs}
\label{sec:pilot}
\ls We begin with a pilot experiment that confirms current PLMs' responses to FPQs are not satisfactory despite their knowledge. We query the PLMs with the questions taken from \OURS test split (see Section~\ref{sec:dataset}). We use the large PLMs whose API is publicly available, including Bloom~\cite{scao2022bloom}, OPT~\cite{zhang2022opt}, Jurassic-1~\cite{lieber2021jurassic}, GPT-3(text-davinci-003)~\cite{brown2020language} (as known as InstructGPT). We use the prompt ``\textit{Question: \makebox[6mm]{\hrulefill} Answer:}'', where the blank is filled by the question text. We provide the generated answers of these models in Table~\ref{tab:direct_answer_examples_main}. We also provide our model's answer (See Section~\ref{sec:exp}) as comparisons. As we can see, all models fail on these simple FPQs. However, in the column ``Ablation'',  we are surprised to find that all models give the correct responses to the questions that ask directly about the correctness of the premises. This motivates us to hypothesize that the inability of current PLMs to handle FPQs is due to distribution mismatch, instead of missing knowledge. Therefore, we need a dataset specializing in  FPQs.

\section{Dataset}
\label{sec:dataset}
To build a dataset on FPQs, there are potentially two approaches. An approach is to collect them from natural corpora. However, false premise questions rarely appear in natural corpora, which makes the question collection process laborious. Second, even if we collect false premise questions, the false premises are made by humans and thus are hard to be detected by humans, which doesn't fit with the motivation of this paper. In fact, ~\citet{min2022crepe} have done pioneering work using this approach. On the contrary, our approach is to manually write such false premise questions. To ensure the quality of our dataset, we expect \OURS dataset to have the following key features: \textit{broad coverage}, \textit{high quality}, \textit{few shortcuts}, and \textit{detailed explanations} for the false premises. Below we introduce the annotation steps that ensure these features.

\subsection{Categorization of FPQs.} 
\ls People ask questions in a wide variety of contexts and formats. Increasing the coverage of questions is proven to be beneficial~\cite{khashabi-etal-2020-unifiedqa}. However, asking annotators to write FPQs freely does not guarantee the coverage of the questions. Therefore, the authors manually think up 29 initial FPQs (see Appendix~\ref{app:intial_fpqs}). Then we categorize these FPQs in terms of error types, and question format. We summarize the categories in Table~\ref{tab:error_types}. 
In total, there are eight error types covering commonsense errors, logical errors, etc., and six question formats covering factual questions, descriptive questions, etc. Although we try to collect as many examples as possible into the initial set, the categorizations are far from exhaustive.  Therefore we include an ``Others'' option to encourage creativity.

\ls\textbf{Writing FPQs.} We recruit twenty human annotators to think up questions that contain false premises. To make the creative process easier, we provide source words to the annotators to compose sentences. We use the subject word of GenericsKB~\cite{bhakthavatsalam2020genericskb} as the source word since they have broad coverage and each word is paired with a short illustrative sentence that can also inspire the annotators.  However, we don't require the annotated sentence to contain the source word. Moreover, the annotators have the freedom to skip the source words that are not easy to brainstorm. We then ask the annotators to categorize the questions into the above categories. The annotators are required to keep a balanced distribution (see Appendix~\ref{app:distribution_balance}) over categories when they finish their part. For the quality of the written FPQs, we require them to be correct in syntax and contain obvious false premises.

% \begin{figure}[!htbp]
% %  \renewcommand\thefigure{4}
% % \begin{minipage}{7cm}
%     \centering
%     \includegraphics[width=1.0\linewidth]{figs/fpqa_catagories.pdf}
%     \caption{{\todo{The} fraction of sub-categories in FPQ datasets from three perspectives. Sectors are decreasing in size along the counterclockwise direction.} }
%     \label{fig:fpqa_catagories}
% \end{figure}

\vspace{-0.7em}
\begin{figure}[!htbp]
    \centering
    \includegraphics[width=\linewidth]{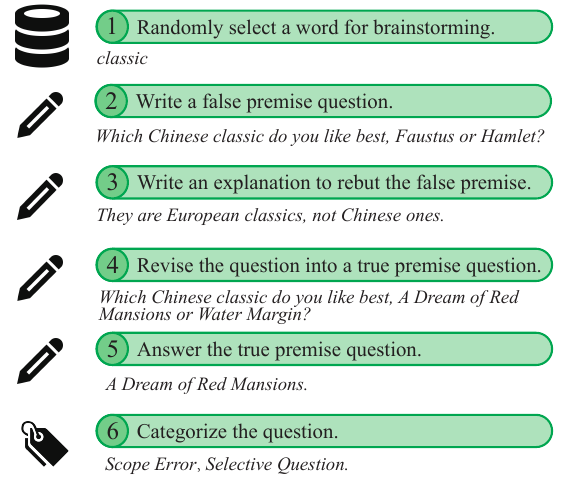}
    \caption{The annotation process of \OURSNOSPACE. The \textit{italic} sentences are one annotation example.}
    \label{fig:annotateprocess}
\end{figure}
\vspace{-0.5em}

\textbf{Revising into TPQs.} Previous studies~\cite{du-etal-2021-towards} point out that PLMs are skilled at finding shortcuts in datasets and do not really understand the task. Since the FPQs are created manually, it's easy to fall into the fixed writing style of the annotators. To alleviate the problem, we annotate a comparison set for these FPQs. Specifically, we ask annotators to edit each FPQ with minimal modifications to make it a problem with true premises (TPQ). The resulting pairs of questions differ only in the correctness of the premises, ensuring that the model learns the essentials of the task.

\textbf{Writing Detailed Explanations/Answers.} Humans usually reply to FPQs with an explanation of why the premise is false~\cite{kaplan1978indirect}. Generating the explanation also helps check whether the model truly understands the FPQs. Therefore, we ask the annotators to write an explanation for each FPQ. For quality control of the explanations, we require the explanation to be more than the negation of the false premise. For the training set and validation set, we require one explanation per question, for the test set, we require two explanations per question. For symmetry, the annotators also write answers to the TPQs. The full annotating process is demonstrated in Figure~\ref{fig:annotateprocess}.

% \begin{figure}[!htbp]
%     \centering
%     \includegraphics[width=1\linewidth]{figs/errordistribution_crop.png}
    
%     \includegraphics[width=0.73\linewidth]{figs/questiontypedistribution_crop.png}
%     \caption{The distributions of the FPQ datasets. We discourage the annotators from biasing towards specific types of FPQs }
%     \label{fig:distribution}
% \end{figure}

%  

\subsection{Dataset Statistics}

\ls The final dataset, dubbed as FalseQA,  contains 2365 question pairs. A snapshot of the FPQ dataset is in Table~\ref{tab:FalseQA_examples}. We randomly split the dataset into train, validation, and test splits, with a ratio of 5:2:3. The summary of statistics is shown in Table~\ref{tab:statisticsofFPQ}.
 
\definecolor{greycolor}{rgb}{0.917, 0.917, 0.917}
 \newcolumntype{g}{>{\columncolor{greycolor}}c}
\begin{table}[!htbp]
    \centering
\resizebox{\linewidth}{!}{
    \begin{tabular}{g|c}
\toprule
Number of annotators & 20 \\
Number error types (FPQs) & 8 \\
Number question format (FPQs) &  6 \\
Average question length (FPQs)& 10.6 tokens\\
Average explanation length (FPQs) & 12.1 tokens\\
Average question length (TPQs) & 10.4 tokens\\
Average answer length (TPQs) & 9.8  tokens\\
 Training set & 1187 question pairs  \\
 Validation set & 491 question pairs \\
 Test set & 687 question pairs \\
\bottomrule
    \end{tabular}
}
    \caption{Statistics of \OURS dataset. The number of tokens is calculated by NLTK~\cite{bird-loper-2004-nltk}.}
    \label{tab:statisticsofFPQ}
    \vspace{-7mm}
\end{table}

\begin{table*}[!htbp]
    \centering
\resizebox{\textwidth}{!}{
    \begin{tabular}{m{2.5cm}|c|m{7cm}|m{8.5cm}}
\toprule
Source Word & Type & \multicolumn{1}{>{\centering\arraybackslash}c|}{Question} & \multicolumn{1}{>{\centering\arraybackslash}c}{Explanation/Answer} \\
\hline
\multirow{2}{*}{tennis} & FPQ & What was the place where the tennis match was launched in the 1200s?& Modern tennis had not been invented in the 12th century. \\
& TPQ & What was the place where the French Open was held in 2021?      &  The 2021 French Open was held in Roland Garros from May to June. \\
  \midrule
\multirow{2}{*}{software} & FPQ & List a software that is developed by Edison. & Edison was a physics inventor, not a computer scientist. \\
 & TPQ & List a software that is developed by Bill Gates. & Windows xp. \\
\bottomrule
    \end{tabular}
}
\vspace{-2mm}
    \caption{Example question pairs (FPQ and TPQ) and their source words, explanations/answers.}
    \label{tab:FalseQA_examples}
    \vspace{-1em}
\end{table*}

\section{Experiments}
\label{sec:exp}
Our experiments are divided into two main parts. To begin with, we conducted extensive experiments to demonstrate that PLMs have the ability to discriminate and rebut FPQs with moderate training data. Next, we propose a practical method to handle both FPQs and general questions well. 

\subsection{Models and Settings}
\ls PLMs are usually divided into three main architectures, namely, encoder-only, decoder-only, and encoder-decoder language models. Since the encoder-only language model can not be used as the QA model, we select typical PLMs from the latter two for experiments.

\ls For decoder-only models, we choose {OPT}~\cite{zhang2022opt}, which is a series of open-source pre-trained models aligned to OpenAI GPT-3~\cite{brown2020language}. 
% OPT uses a decoder-only architecture and auto-gressive training strategy. We test four model scale, i.e., OPT-350M, OPT-1.3B, OPT-2.7B, OPT-175B~\footnote{Using the \href{https://opt.alpa.ai/\#generation}{API} hosted by \href{https://github.com/alpa-projects/alpa}{Alpa project}.}
For the encoder-decoder models, we use T5~\cite{raffel2020exploring} and \textsc{Macaw}~\cite{tafjord2021general}.  T5~\cite{raffel2020exploring} models are trained with the massive unsupervised pre-training corpus and a mixture of supervised tasks, making them very capable of solving various downstream tasks. \textsc{Macaw} is fine-tuned from T5 models on QA tasks. They achieve state-of-the-art performance on direct QA dataset ARC-DA~\cite{bhakthavatsalam2021think} and perform satisfactorily on most categories of the demanding dataset Challenge300~\cite{tafjord2021general} except for the FPQs.

Unless specified, all experiments are repeated three times with different random seeds. For each result, we report the mean and standard deviation. The detailed hyperparameters for each experiment are in Appendix~\ref{app:expdetail}.

\begin{table}
\resizebox{0.97\linewidth}{!}{
    \begin{tabular}{l|c|c|c}
\toprule
      \multicolumn{1}{>{\centering\arraybackslash}c|}{Model}   & Recall &  Precision & Accuracy \\
\hline
    % Bert-Base & 59.9 $\pm$ 7.1 &  72.0 $\pm$ 1.8 & 68.2 $\pm$ 1.2\\
    % Bert-Large & {53.4$\pm$12.2} &   {76.3$\pm$5.0} &  {68.0$\pm$2.2}  \\
    % OPT-QA-350M & 63.7 $\pm$ 6.1&66.0 $\pm$ 1.5&65.4 $\pm$ 1.7 \\
    % OPT-QA-1.3B & 70.0 $\pm$ 7.2&70.6 $\pm$ 3.3&70.2 $\pm$ 0.3\\
    % OPT-QA-2.7B & 74.4 $\pm$ 7.8&73.1 $\pm$ 3.6&73.2 $\pm$ 0.6 \\
    % T5-Large & 75.9 $\pm$ 9.2&73.2 $\pm$ 6.6&73.5 $\pm$ 2.7 \\
    % T5-3B & 79.7 $\pm$ 12.7&80.1 $\pm$ 7.9&79.1 $\pm$ 1.0 \\
    % \textsc{Macaw}-Large & 62.0 $\pm$ 6.6&80.6 $\pm$ 3.5&73.3 $\pm$ 0.9  \\
    % \textsc{Macaw}-3B & 85.9 $\pm$ 0.6&80.4 $\pm$ 2.3&82.5 $\pm$ 1.4 \\
    % \textsc{Macaw}-11B & 82.8 $\pm$ 8.5&82.9 $\pm$ 0.9&82.9 $\pm$ 3.8 \\

    OPT-350M & \std{64.8}{7.2} &	\std{65.5}{3.3} &	\std{65.1}{1.8} \\
    OPT-1.3B & \std{67.4}{7.6} &	\std{73.5}{5.1} &	\std{71.2}{0.4} \\
    OPT-2.7B & \std{69.2}{12.2}	&\std{76.7}{5.0} & 	\std{73.7}{2.1} \\
    T5-Large & \std{72.8}{2.3} &	\std{76.9}{1.5}	& \std{75.4}{0.3}  \\
    T5-3B & \std{80.6}{7.7}	& \std{83.8}{4.3} &	\std{82.3}{1.9} \\
T5-11B & \std{86.5}{1.7} &	\std{82.4}{1.0} &	\std{84.0}{1.1} \\
    \textsc{Macaw}-Large & \std{75.0}{4.1} & 	\std{77.9}{3.3} &	\std{76.7}{0.7}  \\
    \textsc{Macaw}-3B & \std{79.9}{6.8} &	\std{85.0}{5.3} & 	\std{82.6}{0.5} \\
    \textsc{Macaw}-11B & \std{86.0}{2.1} &	\std{87.0}{0.7}	& \std{86.6}{1.3}\\
\bottomrule
    \end{tabular}
}
\vspace{-2mm}
    \caption{The recall and precision are for discriminating FPQs, and the accuracy of binary classification.}
    \label{tab:binary}
    \vspace{-5mm}
% \end{minipage}
\end{table}

% \noindent\textbf{T0++.} T0++~\cite{sanh2021multitask} is trained to understand human instructions by converting a large set of tasks into a unified generation format with natural language prompts. It compares favorably to GPT3~\cite{} on many tasks. We tested the largest checkpoint which has 11 billion parameters~\footnote{Using web API, e.g.,  \url{https://huggingface.co/bigscience/T0pp?text=How+many+legs+do+my+eyes+have\%3F} }. 

% \noindent\textbf{BlenderBot.} BlenderBot~\cite{roller-etal-2021-recipes} is a large-scale open-domain chatbot that is equipped with multiple conversational skills. We ask questions to BlenderBot in conversational form and treat its reply as the answer.  We tested the largest checkpoint with 9 billion parameters.

% \noindent\textbf{Macaw.} Macaw~\cite{tafjord2021general} is a multi-angle question answering model. We use the direct question-answer mode, which uses template XXX to wrap the questions.
% We tested the largest checkpoint with 11 billion parameters.

% \noindent\textbf{Jurassic-1.} Jurassic-1~\cite{lieber2021jurassic} is a modular reasoning system with a large PLMs as its backbone. It can give reasonable answers to challenging questions by looking at multiple data sources.

% \noindent\textbf{GPT-3.} GPT-3~\cite{brown2020language} is a powerful zero-shot language model with an astonishing ability to generate human-like answers. Although released in 2020, it could use data that humans interact with to continuously improve the model~\footnote{See  \url{https://help.openai.com/en/articles/5722486-how-your-data-is-used-to-improve-model-performance}}, which may explain its strong performance. The API calls to GPT-3 is in Appendix~\ref{app:apicalls}.

\subsection{Discriminating FPQs}
\ls We first train the PLMs to classify the question in \OURS into FPQ and TPQ. To mitigate the gap between pre-training  and fine-tuning, we adopt the prompt learning paradigm~\cite{schick-schutze-2021-exploiting,10.1145/3560815} to do the classification. We report the accuracy of the classification. Besides, we report the recall and precision for FPQs since we emphasize the FPQs.

From Table~\ref{tab:binary}, we can see all the models can achieve non-trivial performance on the binary classification. 
% This proves our conjecture that the PLMs can discriminate the FPQs from TPQs without relying on external assistance such as knowledge bases or fact verification. 
(1) The most powerful model \textsc{Macaw}-11B, can achieve 86.6 accuracy. (2) Across all the models of the same type, performance boosts when the size of the model increases. We hypothesize that the scaling effect is because larger models both contain more knowledge and are easier to be activated to understand the task. (3) There is a slight improvement from T5 to \textsc{Macaw}, showing that the ability to identify FPQs can be enhanced by fine-tuning on a corpus of normal questions. 

\subsection{Impact of Training Data Size}
\begin{figure}
    \centering
    \includegraphics[width=0.9\linewidth]{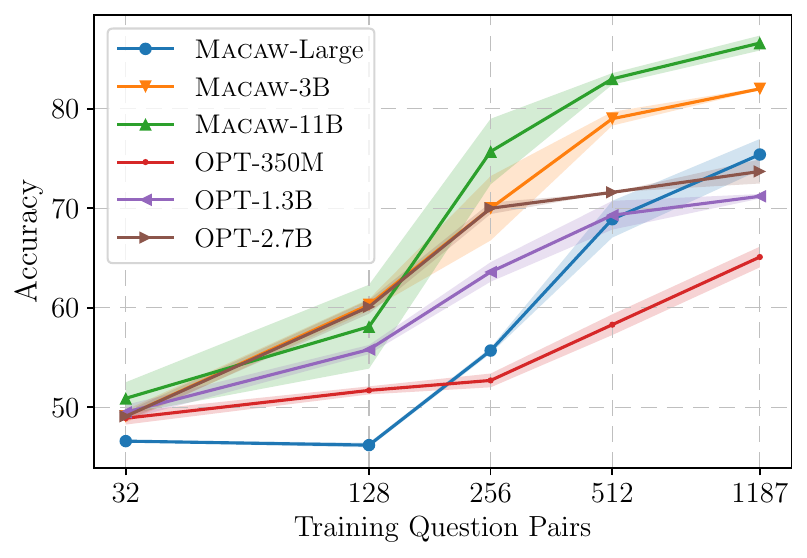}
    \vspace{-0.3cm}
    \caption{PLMs discrimination ability to FPQs from TPQs with the number of training samples. }
    \label{fig:scale}
    \vspace{-0.7cm}
\end{figure}

\ls Then we study the PLMs' performance to discriminate FPQs with fewer training data. We randomly sample 32, 128, 256, and 512 pairs of FPQ and TPQs as the training data and plot the performance under each data scale in Figure~\ref{fig:scale}. We can see that the accuracy of classifying FPQs and TPQs grows almost linearly as the number of pairs grows exponentially. With only 256 pairs of questions, models larger than 2.7B, i.e., OPT-2.7B, \textsc{Macaw}-3B, \textsc{Macaw}-11B, all achieve more than 70\% accuracy, while the smaller models need more data to achieve non-trivial performance. The trade-off between model scale and data scale hints that larger models might be activated with even fewer training data.
However, as we have noticed, the gap between human performance and model performance remains large, as an average person can almost completely classify such problems.

\ls The above results already allow us to design a primitive QA pipeline that can handle FPQs. For example, if the model predicts that a question is FPQ, then it  refuses to answer such questions, while for other questions it generates the answer.

\begin{table}[htbp]
    \centering
\resizebox{\linewidth}{!}{
    \begin{tabular}{c|l|c|c|c}
\toprule
{\# QP} & \multicolumn{1}{>{\centering\arraybackslash}c|}{Model}  & Recall & Accuracy & \textsc{Rouge}-L  \\
\hline
\multirow{6}{*}{32} 
& OPT-2.7B & \myemph{\std{62.4}{14.0}} & \std{52.8}{0.7} & \myemph{\std{27.7}{1.9}}  \\
% \std{62.4}{14.0}&\std{52.8}{0.7}&\std{27.7}{1.9}
& {\quad{+Binary Loss}} & \std{59.0}{5.3} & \myemph{\std{56.3}{1.2}} & \std{27.0}{1.6}\\
& \textsc{Macaw}-3B  & \myemph{\std{41.9}{22.3}} & \std{56.8}{3.4} & \std{29.1}{3.0}  \\
% \std{41.9}{22.3}&\std{56.8}{3.4}&\std{29.1}{3.0}
& {\quad{+Binary Loss}} & \std{40.5}{21.8} & \myemph{\std{61.5}{7.7}} & \myemph{\std{32.0}{1.3}}\\
& \textsc{Macaw}-11B & \myemph{\std{64.5}{36.9}} &  \std{59.2}{9.0} & \myemph{\std{36.2}{5.2}} \\
% \std{64.5}{36.9}&\std{59.2}{9.0}&\std{36.2}{5.2}
& {\quad{+Binary Loss}} & \std{49.0}{19.6} & \myemph{\std{64.1}{7.2}} & \std{33.8}{0.5}\\
\midrule
\multirow{6}{*}{256} 
& OPT-2.7B & \std{56.8}{5.3} & \std{56.9}{2.0} & \std{29.5}{0.4} \\
% \std{56.8}{5.3}&\std{56.9}{2.0}&\std{29.5}{0.4}
& {\quad{+Binary Loss}} & \myemph{\std{62.5}{5.5}} & \myemph{\std{67.8}{1.6}} & \myemph{\std{29.7}{0.5}}  \\
% \std{62.5}{5.5}&\std{67.8}{1.6}&\std{29.7}{0.5}
&\textsc{Macaw}-3B  &  \std{69.5}{7.5}&\std{73.5}{1.7}&\std{34.5}{1.3}  \\
% \std{69.5}{7.5}&\std{73.5}{1.7}&\std{34.5}{1.3}
& {\quad{+Binary Loss}} &  \myemph{\std{72.6}{8.7}} & \myemph{\std{76.5}{2.3}} & \myemph{\std{35.3}{1.5}} \\
% \std{72.6}{8.7}&\std{76.5}{2.3}&\std{35.3}{1.5}
& \textsc{Macaw}-11B  & \std{77.3}{13.0} & \std{76.2}{1.9} & \std{35.0}{2.0}  \\
% \std{77.3}{13.0}&\std{76.2}{1.9}&\std{35.0}{2.0}
& {\quad{+Binary Loss}} & \myemph{\std{81.3}{4.6}} & \myemph{\std{79.2}{0.2}} & \myemph{\std{38.4}{0.7}} \\
% \std{81.3}{4.6}&\std{79.2}{0.2}&\std{38.4}{0.7}
\midrule
\multirow{6}{*}{1187} 
& OPT-2.7B & \myemph{\std{76.2}{4.1}} & \std{70.8}{0.9} & \myemph{\std{34.2}{0.6}} \\
% \std{76.2}{4.1}&\std{70.8}{0.9}&\std{34.2}{0.6}
& {\quad{+Binary Loss}} & \std{75.9}{4.9} & \myemph{\std{75.3}{0.5}} & \std{34.0}{1.1} \\
% \std{75.9}{4.9}&\std{75.3}{0.5}&\std{34.0}{1.1}
& \textsc{Macaw}-3B  & \myemph{\std{81.8}{7.3}} & \std{80.6}{1.2} & \myemph{\std{39.2}{1.9}}  \\
% \std{81.8}{7.3}&\std{80.6}{1.2}&\std{39.2}{1.9}
& {\quad{+Binary Loss}} & \std{80.9}{1.2} & \myemph{\std{84.2}{0.7}} & \std{38.1}{1.0} \\
% \std{80.9}{1.2}&\std{84.2}{0.7}&\std{38.1}{1.0}
& \textsc{Macaw}-11B  & \myemph{\std{90.7}{5.2}} &  \std{83.6}{0.8} & \std{41.9}{0.6}  \\
% \std{90.7}{5.2}&\std{83.6}{0.8}&\std{41.9}{0.6}
& {\quad{+Binary Loss}} & \std{88.8}{1.8} & \myemph{\std{87.1}{0.9}} & \myemph{\std{42.0}{0.7}} \\
% \std{88.8}{1.8}&\std{87.1}{0.9}&\std{42.0}{0.7}
\bottomrule
    \end{tabular}
}
    \caption{Joint FPQ discrimination and explanation generation. Better results are shown in \colorbox{emphcolor}{green}.}
    \label{tab:model_stimulation}
    \vspace{-1.5em}
\end{table}

\subsection{Answering FPQs with Explanations}
\label{exp:answerfpq}
Next, we train the PLMs to discriminate and generate explanations for the FPQs at the same time. 
% The purpose for this experiment is twofold. First, generating the explanation may rule out the possibility that the models are just finding shortcuts in our datasets. Secondly, the explanation serves as the rebuttal to the question which is aligned to human reactions to FPQs.
Since we need to start from models that already have zero-shot QA ability, we choose only \textsc{Macaw} for the encoder-decoder models. For the decoder-only model, we follow similar approaches to \citet{tafjord2021general} to train OPT models with a fraction of UnifiedQA dataset~\cite{khashabi-etal-2020-unifiedqa} in order to steer the model into QA mode~\footnote{We will release the checkpoint.} without injecting much additional knowledge. We select the model size that can achieve non-trivial performance using 256 pairs of data for this experiment.

\ls To discriminate and generate explanations jointly, we let the models generate the discriminating tokens: ``\textit{tricky question}'' or ``\textit{true question}'' first. Then the model continues to generate the explanation to FPQs or the answer to TPQs. Since the numbers of tokens responsible for discrimination and generation differ dramatically, we add an additional binary loss on the discriminating tokens. The ratio between the binary loss and the generation loss is 1.  We conduct experiments on three training data sizes, i.e, 32, 256, and 1187 question pairs.

In evaluation, if a generated answer contains ``\textit{tricky question}'', we consider the question classified as an FPQ, otherwise, it is classified as a TPQ. Similar to the previous section, we report the recall, precision of predicting FPQs, and accuracy of the binary classification. In addition, we evaluate the quality of the generated explanation by computing the maximum \textsc{Rouge}-L~\cite{lin2004rouge} score between it and the two ground-truth explanations. Note since we focus on the explanation of FPQs, the evaluation does not include the TPQs.

From Table~\ref{tab:model_stimulation}, we have three observations. (1) The models jointly predict the question and generate answers successfully. (2) When training data is limited, e.g., 32 question pairs, the accuracy is significantly higher than conducting classification alone (See in Figure~\ref{fig:scale}), which shows that the explanations of the FPQs help the model to quickly adapt to the task.  (3) Adding binary loss boosts the model's performance on classification. For the generated explanations, the best \textsc{Rouge}-L achieves 42.0, showing that the explanations are close to humans'. The quality of explanations also gets higher as the model size and data size increase. We provide the model-generated explanation for 10 randomly sampled FPQs in Appendix~\ref{app:samples}. We can see the explanations are reasonable.

\begin{table}[htbp]
    \centering
\resizebox{0.97\linewidth}{!}{
    \begin{tabular}{c|c|c|c|c}
\toprule
{\# QP} & {Model}  & Recall & Accuracy & \textsc{Rouge}-L  \\
% \multicolumn{5}{c}{\cellcolor[HTML]{D0D0D0}{\textbf{In-context Learning}}}\\
\hline
\multirow{4}{*}{0}& OPT-66B & 6.8 & 25.8 & 12.2 \\
& Jurassic-1  & 66.2 & 36.5 & 6.5 \\
& GPT-3(001) & 46.9 & 46.1 & 5.1\\
& GPT-3(002)  & \myemph{98.5} & \myemph{53.2}  & 
\myemph{25.3} \\
\midrule
\multirow{4}{*}{2}& OPT-66B & \std{21.3}{18.5}&\std{53.0}{2.6}&\std{32.2}{2.8}\\
% \std{21.3}{18.5}&\std{53.0}{2.6}&\std{32.2}{2.8}
& Jurassic-1  & \std{52.8}{37.0}&\std{56.9}{2.6}&\std{32.4}{5.3}\\
% \std{52.8}{37.0}&\std{56.9}{2.6}&\std{32.4}{5.3}
& GPT-3(001) & \std{43.6}{16.7}&\std{63.9}{4.1}&\std{31.8}{2.7}\\
% \std{43.6}{16.7}&\std{63.9}{4.1}&\std{31.8}{2.7}
& GPT-3(002)  & \myemph{\std{87.9}{2.4}} & \myemph{\std{75.2}{1.6}} & \myemph{\std{38.1}{1.5}}\\
% \std{87.9}{2.4}&\std{75.2}{1.6}&\std{38.1}{1.5}
\midrule
\multirow{4}{*}{4} & OPT-66B & \std{19.7}{29.8}&\std{51.9}{3.7}&\std{34.8}{1.4}\\
% \std{19.7}{29.8}&\std{51.9}{3.7}&\std{34.8}{1.4}
& Jurassic-1  & \myemph{\std{94.7}{8.2}}&\std{53.1}{4.8}&\std{38.4}{0.7} \\
% \std{94.7}{8.2}&\std{53.1}{4.8}&\std{38.4}{0.7}
& GPT-3(001)  & \std{61.9}{15.7} &\std{67.6}{1.5}&\std{34.5}{1.2}\\
% \std{61.9}{15.7}&\std{67.6}{1.5}&\std{34.5}{1.2}
& GPT-3(002)  & {\std{90.6}{4.6}} &\myemph{\std{75.8}{2.9}}&\myemph{\std{39.1}{1.6}}\\
% \std{90.6}{4.6}&\std{75.8}{2.9}&\std{39.1}{1.6}
\bottomrule
    \end{tabular}
}
    \caption{Performance of in-context learning under different numbers of examples. Better results are in \colorbox{emphcolor}{green}.}
    \label{tab:incontextlearning}
    \vspace{-1em}
\end{table}
\begin{figure*}[htbp]
    \centering
    \includegraphics[width=0.97\textwidth]{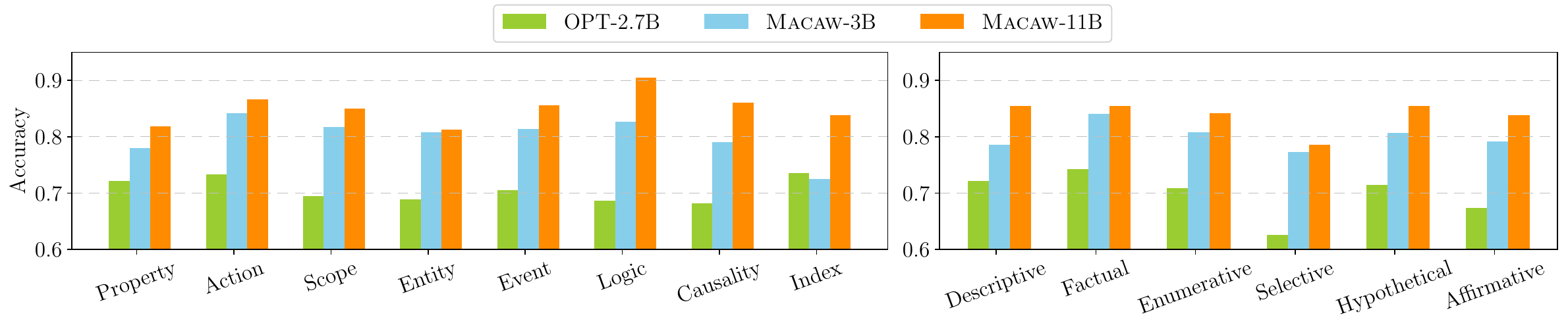}
    \vspace{-2mm}
    \caption{PLMs' accuracy scores for different error types (left) and question formats (right).}
    \label{fig:category}
    \vspace{-1mm}
\end{figure*}

\begin{table*}[!htbp]
    \centering
\resizebox{0.90\linewidth}{!}{
\vspace{-5mm}
    \begin{tabular}{l|cccc|ccc}
\toprule
 \multirow{2}{*}{Settings} & \multicolumn{4}{c|}{FalseQA} & \multicolumn{3}{c}{ARC-DA} \\
\cline{2-5} \cline{6-8} 
 & Recall & Precision & Accuracy & \textsc{Rouge}-L & FPR($\downarrow$) & \textsc{Rouge}-L & F1 \\
\midrule
       Raw \textsc{Macaw}-11B  & \std{8.7}{2.5}&\std{91.5}{7.8}&\std{53.8}{0.8}&\std{7.2}{0.0}&\std{0.0}{0.0}&\std{54.5}{0.0}&\std{55.0}{0.0} \\
% \std{8.7}{2.5}&\std{91.5}{7.8}&\std{53.8}{0.8}&\std{7.2}{0.0}&\std{0.0}{0.0}&\std{54.5}{0.0}&\std{55.0}{0.0}
\midrule
    \quad + FPQ (256 shots) &
    \myemph{\std{81.3}{4.6}}&\std{78.2}{2.2}&\myemph{\std{79.2}{0.2}}&\myemph{\std{38.4}{0.7}}&\std{23.9}{13.6}&\std{24.2}{1.5}&\std{23.9}{1.6}\\
% \std{81.3}{4.6}&\std{78.2}{2.2}&\std{79.2}{0.2}&\std{38.4}{0.7}&\std{23.9}{13.6}&\std{24.2}{1.5}&\std{23.9}{1.6}\\
      \quad\qquad + Data Replay & \std{72.1}{7.0}&\myemph{\std{81.4}{0.9}}&\std{77.9}{3.1}&\std{35.1}{1.0}&\myemph{\std{1.8}{0.9}}&\myemph{\std{30.6}{2.9}}&\myemph{\std{30.4}{3.0}} \\
% \std{72.1}{7.0}&\std{81.4}{0.9}&\std{77.9}{3.1}&\std{35.1}{1.0}&\std{1.8}{0.9}&\std{30.6}{2.9}&\std{30.4}{3.0}\\
\midrule
\quad+ FPQ (Full)  &
\myemph{\std{88.8}{1.8}}&\std{85.9}{2.7}&\myemph{\std{87.1}{0.9}}&\myemph{\std{42.0}{0.7}}&\std{12.6}{6.6}&\std{32.2}{2.4}&\std{32.3}{2.5}\\
% \std{88.8}{1.8}&\std{85.9}{2.7}&\std{87.1}{0.9}&\std{42.0}{0.7}&\std{12.6}{6.6}&\std{32.2}{2.4}&\std{32.3}{2.5}\
\quad\qquad + Data Replay & \std{85.6}{1.3}&\myemph{\std{87.5}{0.5}}&\std{86.7}{0.5}&\std{39.2}{0.8}&\myemph{\std{1.4}{0.0}}&\myemph{\std{48.6}{1.4}}&\myemph{\std{49.1}{1.2}} \\
% \std{85.6}{1.3}&\std{87.5}{0.5}&\std{86.7}{0.5}&\std{39.2}{0.8}&\std{1.4}{0.0}&\std{48.6}{1.4}&\std{49.1}{1.2}\\
\midrule
       Raw OPT-2.7B  & \std{5.0}{2.0}&\std{54.5}{14.8}&\std{50.5}{1.3}&\std{7.3}{0.0}&\std{0.1}{0.0}&\std{39.4}{0.0}&\std{39.0}{0.0} \\
\midrule
    \quad + FPQ (256 shots) & 
   \std{62.5}{5.5}&\myemph{\std{70.0}{1.9}}&\std{67.8}{1.6}&\myemph{\std{29.7}{0.5}}&\std{19.9}{3.8}&\std{25.0}{0.2}&\std{23.9}{0.3}\\
% \std{62.5}{5.5}&\std{70.0}{1.9}&\std{67.8}{1.6}&\std{29.7}{0.5}&\std{19.9}{3.8}&\std{25.0}{0.2}&\std{23.9}{0.3}\\
      \quad\qquad + Data Replay & \myemph{\std{64.0}{2.8}}&\std{69.4}{1.0}&\myemph{\std{67.9}{0.4}}&\std{29.1}{1.3}&\myemph{\std{1.8}{0.8}}&\myemph{\std{33.8}{0.7}}&\myemph{\std{33.1}{0.9}} \\
% \std{64.0}{2.8}&\std{69.4}{1.0}&\std{67.9}{0.4}&\std{29.1}{1.3}&\std{1.8}{0.8}&\std{33.8}{0.7}&\std{33.1}{0.9}\\
\midrule
\quad+ FPQ (Full)  &
\std{75.9}{4.9}&\myemph{\std{75.2}{3.0}}&\myemph{\std{75.3}{0.5}}&\myemph{\std{34.0}{1.1}}&\std{33.2}{6.0}&\std{22.0}{0.8}&\std{20.8}{0.9}\\
% \std{75.9}{4.9}&\std{75.2}{3.0}&\std{75.3}{0.5}&\std{34.0}{1.1}&\std{33.2}{6.0}&\std{22.0}{0.8}&\std{20.8}{0.9}\\
\quad\qquad + Data Replay & \myemph{\std{76.8}{2.5}}&\std{74.2}{1.2}&\std{75.0}{0.4}&\std{33.2}{0.5}&\myemph{\std{3.5}{0.3}}&\myemph{\std{35.8}{0.9}}&\myemph{\std{35.3}{1.1}}\\
% \std{76.8}{2.5}&\std{74.2}{1.2}&\std{75.0}{0.4}&\std{33.2}{0.5}&\std{3.5}{0.3}&\std{35.8}{0.9}&\std{35.3}{1.1}\\
\bottomrule
    \end{tabular}
}
\vspace{-2mm}
    \caption{Results after tuning with \OURS data and data replay techniques.  Better results are shown in \colorbox{emphcolor}{green}.}
    \label{tab:replay}
    \vspace{-6mm}
\end{table*}

\vspace{-2mm}
\subsection{In-context Learning}
\vspace{-2mm}
We proceed to study the performance of larger models, e.g., GPT-3(175B) on \OURSNOSPACE. The large PLMs are tuned by in-context learning with frozen model parameters. We select {OPT-66B}~\cite{zhang2022opt}, Jurassic-1~\cite{lieber2021jurassic}, and GPT-3(001) and GPT-3(002)~\footnote{text-davinci-001, and text-davinci-002 checkpoints.}. We present the results in Table~\ref{tab:incontextlearning}. We can see that OPT-66B and Jurassic-1 perform poorly. Therefore, we conclude that due to the distribution mismatch of FPQs to normal questions, it is still hard to activate the rebuttal ability using a few examples for these models, which we leave to future work. GPT-3 can be activated with 2 or 4 pairs of examples, however, its performance is lower than the much smaller fine-tuned models in Section~\ref{exp:answerfpq}. Surprisingly, GPT-3(002) has far better performance than GPT-3(001). We hypothesize that they more easily understand the rebuttal task  since they are trained with instruction tuning~\cite{ouyang2022training}. 
% Since the training data of GPT-3(002) model is not open-sourced, we hypothesis that they might use similar tricky questions from the users or the annotators to enhance the capability.

\vspace{-1mm}
\subsection{Performance w.r.t. Category}
\vspace{-2mm}
To better understand which kind of FPQs is harder to be discriminated against, we draw the accuracy of each category in Figure~\ref{fig:category}. In spite of the inconsistency between PLMs, index error is generally hard to classify while logic and causality error is easy. For question types, selective questions are hard to classify while factual questions are easy. These observations can guide the future improvement of our dataset.

\subsection{Answering FPQs and General Questions }
% In our construction of the FPQ dataset, we provide TPQs as comparators for FPQs. However, the distribution of our TPQs may be different from the data distribution to which the model was originally adapted.
QA models are originally used to answer general questions, e.g., questions in ARC-DA~\cite{bhakthavatsalam2021think}~\footnote{Short for AI2 Reasoning
Challenge-Direct Answer.} dataset where the distribution is different from \OURSNOSPACE. Therefore, training purely on \OURS may lead to catastrophic forgetting. To produce a model that handles both FPQs and general questions, we use a simple data replay technique (DR)~\cite{chaudhry2019tiny}. Specifically, during training on \OURS dataset, for each iteration over batches, we add a batch of the data samples from the ARC-DA.  In order to use as little ARC-DA data as possible, we keep the ARC-DA samples to be the same within 30 batch iterations. The aforementioned binary loss is used no matter with or without DR. The concrete numbers of general questions used in each setting and training details are in Appendix~\ref{app:replay}.

In Table~\ref{tab:replay}, we summarize the performance of the raw model before training on \OURSNOSPACE, the model tuned on \OURSNOSPACE, and the model tuned on \OURS with DR. For the original models, since they do not generate the ``\textit{tricky question}'' or ``\textit{true question}'', we manually read the generated answers for 100 randomly sampled questions pairs to determine whether it contains any rebuttals. As we can see, before fine-tuning on FPQs, the models perform well on the ARC-DA dataset. However, they fail substantially on \OURSNOSPACE. After tuning on \OURSNOSPACE, though the models' rebuttal ability is activated, \textsc{Rouge}-L and F1 scores on ARC-DA drop considerably. The false prediction rate (FPR), i.e., the fraction of ARC-DA questions that are incorrectly labeled as tricky questions, is non-negligible. Fortunately, when we apply the DR technique, models not only have small FPRs and the improved quality of generated answers on ARC-DA but the same or even better performance on \OURSNOSPACE. We also find the questions in ARC-DA that PLM still rebuts (see Appendix~\ref{app:stillrebuts}) are also reasonable to rebut for humans. The result gives us a promising direction for building QA systems that perform well on general questions and FPQs.

\section{Conclusion}
\ls In this paper, we investigate using PLMs to answer FPQs, which are simple for humans but deceive most PLMs. We present the first human-written dataset of FPQs. Using the dataset, we successfully activate the discrimination and explanation ability of PLMs and produce PLMs that are both capable of general questions and robust to FPQs.  For future directions, we think that more advanced techniques can be used together with \OURS to fully activate the model's ability, e.g., reinforcement learning with human feedbacks~\cite{ouyang2022training}. Incorporating more knowledge into PLMs is also beneficial for PLMs to answer FPQs. 

% \FloatBarrier 

\section*{Acknowledgement}
This work is supported by the National Key R\&D Program of China (No. 2020AAA0106502), China Postdoctoral Science Foundation (No. 2022M721829), and Institute Guo Qiang at Tsinghua University.

\section*{Limitations}
\ls There are several limitations  in our work. (1) Although we think that PLMs' rebuttal ability is activated in our experiments, the performance has a large space for improvement. For a binary classification problem, the most powerful PLM in our experiment reaches 87.1\% accuracy at most. (2) Since it's hard to probe what the PLMs \textit{truly know}, we didn't further investigate whether PLMs still fail  on some FPQs due to a lack of relevant knowledge or other reasons.  (3) A third limitation is that we notice that the newly announced model ChatGPT~\cite{chatgpt} handles such questions satisfactorily. However, since their training data and details are not open-sourced, we are unable to investigate how the ability of these particular models is activated. (4) In this paper, we standardize the expected responses to FPQs as rebuttals, which takes a conventional perspective. However, sometimes we can react with a more creative response, such as a rhetorical question. This can be future work.

\section*{Ethical Statement}
\ls In the construction of the dataset, we forbid the annotators to compose any sentence that is offensive, harmful, or contains personal information. The annotated data is manually checked to ensure safety.  We pay our annotators a competitive salary relative to market rates. The annotated dataset is helpful to encourage models ``think'' before they provide a response, thus being safer in practical deployment.

% \clearpage

% \footnotetext{Using [random.randint(1,3010) for i in range(3)] with random seed 42.}

\bibliography{anthology,custom}
\bibliographystyle{acl_natbib}

\appendix

\clearpage
\section*{Appendices}
\label{sec:appendix}

\section{Annotation Details}
\subsection{Initial FPQs}
\label{app:intial_fpqs}
We provide the annotators with 29 FPQs in the annotation guide. These questions are original references provided for annotators to brainstorm questions. We list the questions and their error types in Table~\ref{tab:initial_fpqs}. We didn't provide FPQs for each question format since the question format is much easier to determine without examples.

% We show the initial 29 FPQs and their category in our annotating guide in Table~\ref{tab:initial_fpqs}. These questions are original references provided for annotators to brainstorm questions and categorize questions they made.
% 标注手册里的FPQs
\begin{table}[htbp]
\vspace{-0.5em}
    \centering
\resizebox{\linewidth}{!}{
    \begin{tabular}{m{1.5cm}|m{7.5cm}}
    \toprule
     \multicolumn{1}{>{\centering\arraybackslash}l|}{\textbf{Error Type}} & \multicolumn{1}{>{\centering\arraybackslash}c}{\textbf{Question}}\\
     \hline
     \multirow{6}{*}{Entity} & What color is the sun's eye?\\
    & Who was the founding president of Qing Dynasty?\\
    & What color are people's feathers?\\
    & Are the bananas on the apple tree delicious?\\
    & Is hydrogen in oxygen combustible?\\
    \midrule
     \multirow{2}{*}{Index} & What is the 50th largest province in China?\\
     & What day is the eighth day of a week?\\
     \midrule
     \multirow{8}{*}{Action} & Why can't the car be parked in the parking lot?\\
     & When did we cure AIDS?\\
     & When did man go to Uranus?\\
     & How does a cat use a computer?\\
     & How high can a dog climb a tree?\\
     & How far a fish can walk?\\
     & How do pupils go to school with their wives?\\
     & How to pry open the walnut plasticine?\\
     \midrule
     \multirow{1}{*}{Property} & How long does it take for the sun to become transparent?\\
     \midrule
     \multirow{5}{*}{Scope} & How do I take the train at the airport?\\
     & What kind of turtle is a fish?\\
     & What causes Oda Nobunaga's death in the Odyssey?\\
     & Who caused Guan Yu's death in Water Margin?\\
     \midrule
     \multirow{5}{*}{Causality} & Why the more water you drink, the more thirsty you become?\\
    & Why is the table in a pen?\\
    & Where is the computer on the motherboard?\\
    & What percentage of California is the United States of America?\\
    \midrule
    \multirow{2}{*}{Logic} & How to sit or stand at the same time?\\
    & Where will the dead come back to life?\\
    \midrule
    \multirow{3}{*}{Event} & How many times did Aristotle use a computer?\\
    & When did Zuckerberg start Google?\\
    & When Homer wrote The Odyssey?\\
     \bottomrule
    \end{tabular}
}
    \caption{Initial FPQs}
    \vspace{-0.5em}
    \label{tab:initial_fpqs}
\end{table}

\subsection{Distribution Balance Criterion}
\label{app:distribution_balance}
% As FPQs can be categorized into error types, question formats, a balanced distribution is required to get an objective analysis result. For eight error types, every type of FPQs should at least take 5\% of the whole data, and the maximum category should be not more than 30\%. And for six question formats, every type of FPQs should at least take 10\% of the whole data, and the maximum category should be not more than 30\%. All of the distribution balance criteria don't take into account the ``Others'' category.
We expect our dataset to have a richer and more uniform distribution of FPQs. We achieve this goal with the help of constraints on the FPQ types. For the eight error types, each type of FPQ should account for at least 5\% of the overall data, and the maximum category should not exceed 30\%. And for the six problem formats, each type of FPQ should account for at least 10\% of the entire data, and the maximum category should not exceed 30\%. All balance criteria do not take into account the ``other" category.

% \subsection{Dataset Filtering}
% We initially annotate 3000 FPQs and 3000 TPQs and their explanation and answer, respectively. We manually checked the whole dataset and screened out the question pairs that don't have obvious false premises or don't have meaningful answers, resulting in 2365 question pairs. We annotate answers to the test split's FPQs again to get more unbiased quality evaluation scores.
% % 验收的分布标准

\section{Experiment Details~\footnote{We choose random seeds $4$, $13$, and $34$ in all experiments.}}
\label{app:expdetail}
\subsection{API Calls for Pilot Experiments}
We summarize the APIs used in Section~\ref{sec:pilot} in Table~\ref{tab:apis}. We will also provide the screenshot of using these APIs in our final reproducible code.

% 每个用到API的url，参数填在下表中

\begin{table*}[!htbp]
\vspace{0cm}
\centering
\resizebox{\textwidth}{!}{
    \begin{tabular}{c|c|c|c}
\toprule 
  Model  & API URL & Prompt Template & Hyperparameters\\
 \hline
Bloom & \url{https://huggingface.co/bigscience/bloom} & Question: \makebox[6mm]{\hrulefill}  Answer: & {Sampling Strategy: greedy} \\
 \midrule
OPT & \url{https://opt.alpa.ai} & Question: \makebox[6mm]{\hrulefill}  Answer: & \makecell{Response Length: 64; \\ Temperature: 0.7; Top-p: 0.7} \\
 \midrule
GPT-3 & \url{https://beta.openai.com/playground} & Question: \makebox[6mm]{\hrulefill}  Answer: &\makecell{ Temperature: 0.7;\\ Maximum length: 256; Top-p: 1} \\
 \midrule
Jurassic-1 & \url{https://api.ai21.com/studio/v1/j1-jumbo/complete} & Question: \makebox[6mm]{\hrulefill} Answer: & \makecell{Temperature: 0; \\TopK: 0; TopP: 1; MaxTokens: 32}\\
  \bottomrule
    \end{tabular}
}
 \caption{The APIs and hyperparameters when using the APIs.}
\label{tab:apis}
\vspace{-0.3cm}
\end{table*}

%以下按照每个实验把batchsize，learningrate做个小表给出
\subsection{Details of Discriminating FPQs}
For the experiments in Table~\ref{tab:binary}, we use the prompt learning~\cite{schick-schutze-2021-exploiting} paradigm. We use ``true'' and ``false'' as the label word for FPQ and TPQ, respectively~\footnote{Since our target is to classify whether it has a false premise, we set True for FPQs and False for TPQs.}. For T5 models, following the usage of T5~\cite{raffel2020exploring} in their original paper, we append ``\textit{potential tricky question:}'' to identify the task.  \textsc{Macaw} models are multi-angle QA models,  to use their direct question angle, we follow their paper and use ``\textit{\$answer\$ ; \$question\$ = }'' as the prefix. For OPT models, we train them in a vanilla input-output format. We list the hyper-parameters for each experiment in Table~\ref{tab:discriminate_hp}. For \textsc{Macaw}-11B, we use half-precision acceleration and do not find performance degradation compared to full-precision computation. For the experiment in Figure~\ref{fig:scale}, we use the same input-output format mentioned before. Our hyperparameters used in this section are listed in Table ~\ref{tab:discriminate_hp}.
 
%  in~\cite{tafjord2021general} when fine tuning \textsc{Macaw} models

% Finally, we have these input-output formats below.
\newtcolorbox{mybox}{colback=gray!6!white, colframe=gray!75!black}

% \begin{mybox}
% T5's \textbf{INPUT}:\\
% % \centerline{potential tricky question: FPQ/TPQ}\\
% potential tricky question: FPQ/TPQ\\
% T5's \textbf{OUTPUT}:\\
% % \centerline{true/false}\\
% true/false\\
% $\textsc{Macaw}$'s \textbf{INPUT}:\\
% % \centerline{\$answer\$ ; \$question\$ = FPQ/TPQ}\\
% \$answer\$ ; \$question\$ = FPQ/TPQ\\
% $\textsc{Macaw}$'s \textbf{OUTPUT}:\\
% % \centerline{\$answer\$ = true/false}
% \$answer\$ = true/false
% \label{tab:input-format}
% \end{mybox}

% In discriminating FPQs on limited training question pairs, we adopt the prompt learning paradigm to get better performances and bridge the gap between pretraining data and fine tuning data, allowing PLMs to make fuller use of the pretraining data\todo{~\cite{}}. % need a cite

% We choose ``true'' and ``false'' as our label words and perform the classification procedure by comparing the logits between ``true'' and ``false''. We also use the mentioned input-output formats without anything changed, Prompt-Engineering\todo{~\cite{}} may lead to  better performances.

 % need a cite

\begin{table}[!htbp]
    \centering
\resizebox{\linewidth}{!}{
    \begin{tabular}{c|c|c|c|c}
\toprule
{\# QP} & \multicolumn{1}{>{\centering\arraybackslash}c|}{Model}  & Learning Rate & Batch Size & Epoch\\

\hline
% \multicolumn{4}{c}{\cellcolor[HTML]{D0D0D0}{\textbf{Sequence to Sequence Paradigm}}}  \\
% \midrule
% \multirow{8}{*}{1187} 
% & OPT-350M & $5e-6$ & $32$\\
% & OPT-1.3B & $5e-6$ & $32$\\
% & OPT-2.7B & $5e-6$ & $32$\\
% &T5-Large  &  $3e-4$ & $32$\\
% &T5-3B  &  $1e-4$ & $32$\\
% &\textsc{Macaw}-Large  &  $1e-4$ & $32$\\
% &\textsc{Macaw}-3B  &  $1e-4$ & $32$\\
% & \textsc{Macaw}-11B  & $1e-4$ & $\todo{16}$\\
% \midrule

% \multicolumn{4}{c}{\cellcolor[HTML]{D0D0D0}{\textbf{Prompt Learning Paradigm}}}  \\
% \midrule
\multirow{5}{*}{32}
& OPT-350M & $1e-5$ & $32$ & $5$\\
& OPT-1.3B & $1e-5$ & $32$ & $5$\\
& OPT-2.7B & $1e-5$ & $32$ & $5$\\
&\textsc{Macaw}-Large  &  $2e-5$ & $32$ & $5$\\
&\textsc{Macaw}-3B  & $1e-4$ & $32$ & $5$\\
& \textsc{Macaw}-11B  & $1e-4$ & $32$ & $5$\\
\midrule
\multirow{5}{*}{128} 
& OPT-350M & $1e-5$ & $32$ & $5$\\
& OPT-1.3B & $1e-5$ & $32$ & $5$\\
& OPT-2.7B & $1e-5$ & $32$ & $5$\\
&\textsc{Macaw}-Large  &  $2e-5$ & $32$ & $5$\\
&\textsc{Macaw}-3B  & $1e-4$ & $32$ & $5$\\
& \textsc{Macaw}-11B  & $1e-4$ & $32$ & $5$\\
\midrule
\multirow{5}{*}{256} 
& OPT-350M & $1e-5$ & $32$ & $5$\\
& OPT-1.3B & $1e-5$ & $32$ & $5$\\
& OPT-2.7B & $1e-5$ & $32$ & $5$\\
&\textsc{Macaw}-Large & $1e-4$ & $32$ & $5$\\
&\textsc{Macaw}-3B & $1e-4$ & $32$ & $5$\\
& \textsc{Macaw}-11B  & $1e-4$ & $32$ & $5$\\
\midrule
\multirow{5}{*}{512} 
& OPT-350M & $1e-5$ & $32$ & $5$\\
& OPT-1.3B & $1e-5$ & $32$ & $5$\\
& OPT-2.7B & $1e-5$ & $32$ & $5$\\
&\textsc{Macaw}-Large  &  $1e-4$ & $32$ & $5$\\
&\textsc{Macaw}-3B  &  $1e-4$ & $32$ & $5$\\
& \textsc{Macaw}-11B  & $1e-4$ & $32$ & $5$\\
\midrule
\multirow{8}{*}{1187} 
& OPT-350M & $1e-5$ & $32$ & $5$\\
& OPT-1.3B & $1e-5$ & $32$ & $5$\\
& OPT-2.7B & $1e-5$ & $32$ & $5$\\
& T5-Large & $1e-4$ & $32$ & $5$\\
& T5-3B & $1e-4$ & $32$ & $5$\\
& T5-11B & $1e-4$ & $32$ & $5$\\
&\textsc{Macaw}-Large  &  $1e-4$ & $32$ & $5$\\
&\textsc{Macaw}-3B  &  $1e-4$ & $32$ & $5$\\
& \textsc{Macaw}-11B  & $1e-4$ & $32$ & $5$\\

\bottomrule
    \end{tabular}
}
    \caption{Hyperparameters for discriminating FPQs.}
    \label{tab:discriminate_hp}
    \vspace{-1.3em}
\end{table}

\subsection{Details of Answering FPQs}
Since fine-tuned models in few-shots (e.g. 32 question pairs) sometimes may not generate ``\textit{tricky/true question}'' at the beginning of sentence~\footnote{Some seeds in OPT models sometimes produce ``this is a \textit{tricky question}''.}, and a normal answer hardly has ``\textit{tricky/true question}'' in it, we count whether ``\textit{tricky question}'' or ``\textit{true question}'' appears in outputs for classification evaluation to get the recall, precision, and accuracy scores. When evaluating the generated explanation, we remove ``\textit{tricky question}'' and  ``\textit{true question}''.  We list our hyperparameters used in this section in Table~\ref{tab:answering_hp} and keep them the same when adding the binary loss.

\begin{table}[!htbp]
    \centering
\resizebox{\linewidth}{!}{
    \begin{tabular}{c|c|c|c|c}
\toprule
{\# QP} & \multicolumn{1}{>{\centering\arraybackslash}c|}{Model}  & Learning Rate & Batch Size & Epoch\\
\hline
\multirow{3}{*}{32} 
& OPT-2.7B & $5e-6$ & $8$ & $16$\\
& \textsc{Macaw}-3B  & $3e-5$ & $8$ & $8$\\
& \textsc{Macaw}-11B & $1e-4$ & $4$ & $3$\\
\midrule
\multirow{3}{*}{256} 
& OPT-2.7B & $3e-6$ & $32$ & $12$\\
&\textsc{Macaw}-3B  &  $3e-5$ & $32$ & $8$\\
& \textsc{Macaw}-11B & $2.5e-4$ & $4$ & $3$\\
\midrule
\multirow{3}{*}{1187} 
& OPT-2.7B & $6e-6$ & $32$ & $8$\\
&\textsc{Macaw}-3B  &  $5e-5$ & $16$ & $8$\\
& \textsc{Macaw}-11B  & $1e-4$ & $4$ & $3$\\
\bottomrule
    \end{tabular}
}
    \caption{Hyperparameters for answering FPQs.}
    \label{tab:answering_hp}
    \vspace{-1.3em}
\end{table}

% \subsubsection{Fine-tune Learning}
% \label{app:finetune}
% We fine tune models in a normal teacher-forcing way. When computing the raw model's classification scores, we randomly choose 100 pairs of FPQ\&TPQ questions (i.e. 200 in total) and evaluate them by humans.

\subsection{Details of In-context Learning}
\label{app:incontext}
In-context learning, introduced in GPT-3~\cite{brown2020language}, has been a successful way of adapting extensive language models. In in-context learning, we provide a textual prefix $p$ of the task and one or a few training data samples before sending the input questions. 
% It is a good test bed for the insome ability is inherited in a PLM, since no parameter update is needed. We test OPT-66B, Jurrassic-X, and GPT-3~\footnote{For OPT-66B, we use fp16 acceleration and do not find performance variation.}.
We adopt the QA prefix in the GPT-3 demo for all the PLMs tested. Specifically, the prefix is:
\begin{mybox}
$p$ = \textit{I am a highly intelligent question answering bot. If you ask me a question that is rooted in truth, I will give you the answer. If you ask me a question that is nonsense, trickery, or has no clear answer, I will say ``tricky question.'' first and give the reason, otherwise I will say ``true question.'' first and give the reason.}
\end{mybox}
A few pairs of \OURS samples $\{(q_F^i, a_F^i), (q_T^i, a_T^i)\}$ can be concatenated to the textual instruction. Therefore the full prefix before the input question has the following form:
\begin{mybox}
\vspace{-5mm}
\begin{align*}
& p + \text{Q:} q_{F}^i + \text{A:} a_{F}^i +\text{Q:} q_{T}^i + \\ & \text{A:} a_{T}^i + ...  + \text{Q:} \makebox[6mm]{\hrulefill} +  \text{A:}\\
\end{align*}
\vspace{-15mm}
\end{mybox}
\noindent where $+$ indicates string concatenation, and the input example is filled into the blank.

We list our hyperparameters for in-context learning in Table~\ref{tab:incontext_apis}.

\begin{table*}[!htbp]
\vspace{0cm}
\centering
\resizebox{0.9\textwidth}{!}{
    \begin{tabular}{c|c|c}
\toprule 
  Model  & API URL & Hyperparameters\\
 \hline
GPT-3 & \url{https://beta.openai.com/playground} &\makecell{ Temperature: 0;\\ Top-p: 1; Maximum length: 32} \\
 \midrule
Jurassic-1 & \url{https://api.ai21.com/studio/v1/j1-jumbo/complete} & \makecell{Temperature: 0; \\TopK: 0; TopP: 1; MaxTokens: 32}\\
  \bottomrule
    \end{tabular}
}
 \caption{The APIs and hyperparameters for performing in-context learning.}
\label{tab:incontext_apis}
\vspace{-0.3cm}
\end{table*}

\subsection{Answering FPQs and General Questions}
\label{app:replay}
We list our hyperparameters in this section in Table~\ref{tab:replay_hp}. We count the number of general questions when using the data replay technique in Table~\ref{tab:replay_seen_data}.

\begin{table}[htbp]
    \centering
\resizebox{\linewidth}{!}{
    \begin{tabular}{c|c|c|c|c}
\toprule
{\# QP} & Model  & Learning Rate & Batch Size & Epoch\\
\hline
\multirow{2}{*}{256} 
& OPT-2.7B & $3e-6$ & $32$ & $12$\\
& \textsc{Macaw}-11B  & $2.5e-4$ & $4$ & $3$\\
\midrule
\multirow{2}{*}{1187} 
& OPT-2.7B & $6e-6$ & $32$ & $8$\\
& \textsc{Macaw}-11B  & $1e-4$ & $4$ & $3$\\
\bottomrule
    \end{tabular}
}
    \caption{Hyperparameters for handling both FPQs and general questions.}
    \label{tab:replay_hp}
    % \vspace{-1.3em}
\end{table}

\begin{table}[htbp]
    \centering
\resizebox{0.75\linewidth}{!}{
    \begin{tabular}{c|c|c}
\toprule
{\# QP} & Model  & {\# General Questions}\\
\hline
\multirow{2}{*}{256} 
& OPT-2.7B & $32$\\
& \textsc{Macaw}-11B  & $20$\\
\midrule
\multirow{2}{*}{1187} 
& OPT-2.7B & $96$\\
& \textsc{Macaw}-11B  & $80$\\
\bottomrule
    \end{tabular}
}
    \caption{How many general questions models seen when performing data replay.}
    \label{tab:replay_seen_data}
    \vspace{-1.3em}
\end{table}

\label{app:morepilot}
\begin{table*}[!htbp]
\vspace{0cm}
\centering
\resizebox{\textwidth}{!}{
\begin{tabular}{m{2.2cm}|m{1cm}|m{8cm}|c|m{8cm}}
\toprule 
  Model  & \# Para & \multicolumn{1}{>{\centering\arraybackslash}c|}{Reply} & Correctness & \multicolumn{1}{>{\centering\arraybackslash}c}{Ablation}\\
 \midrule
   \multicolumn{4}{c}{\cellcolor[HTML]{EAEAEA} Why there is just one season in China? } &  \multicolumn{1}{c}{\cellcolor[HTML]{EAEAEA} How many seasons does China have?}\\
 \midrule 
 Bloom & 176B & Because the Chinese government is afraid of the virus spreading.   & \xmark & China has four seasons ...\\
  OPT & 175B & Because there are too many people ... & \xmark & China has four Seasons. Spring, ... \\
  Jurassic-1 & 178B  &  Because China is a big country. & \xmark & China has four distinct seasons.\\
  GPT-3 & 175B & In China, there is only one season because of its geographical location. & \xmark & China has four seasons: spring, summer, autumn, and winter.\\
 \textsc{Macaw} & 11B & It is a landlocked country. & \xmark & Four.\\
  \textsc{Macaw}-FPQ & 11B & China has four seasons, not one. & \cmark & China has four seasons.\\
\midrule 
\multicolumn{4}{c}{\cellcolor[HTML]{EAEAEA} What was the place where the tennis match was launched in the 1200s?} & \multicolumn{1}{c}{\cellcolor[HTML]{EAEAEA}In which year was the tennis invented?}\\
\midrule
 Bloom & 176B &  The first tennis match was played in the 1200s in the town of ... & \xmark & The tennis was invented in the year 1874.\\
 % I'm not sure, but I do know that it was played on grass, hardcourt, and clay.
  OPT & 175B & The tennis match was launched in France. & \xmark & The tennis was invented in the year 1874.\\
  Jurassic-1 & 178B  &  The tennis court was in the palace of Westminster. & \xmark & Tennis was invented in 1874.\\
  GPT-3 & 175B &  The first recorded tennis match was launched in France ... & \xmark & The modern game of tennis is believed to have been invented in 1873...\\
   % by Major Walter Clopton Wingfield.
   \textsc{Macaw} & 11B & Paris. & \xmark & 1887. \\
  \textsc{Macaw}-FPQ & 11B &  Tennis was invented in the 19th century. & \cmark &  In 1887.\\
\midrule 
 \multicolumn{4}{c}{\cellcolor[HTML]{EAEAEA} List a software that is developed by Edison. } & \multicolumn{1}{c}{\cellcolor[HTML]{EAEAEA}Did Edison develop any software?}\\
 \midrule 
Bloom & 176B &   Edison is a software that is developed by Edison. & \xmark & Yes, he invented the first ... \\
  OPT & 175B & The Edison Mail. & \xmark & Yes. Edison invented the first electric light bulb,\\
  Jurassic-1 & 178B  & Edison is developed by Edison.  & \xmark & Yes, Edison developed several pieces of software.\\
  GPT-3 & 175B & Edison Design System (EDS) is a software suite developed by Edison ... & \xmark & No, Thomas Edison did not develop any software... \\
  % He is most famous for developing numerous inventions, such as the phonograph, the light bulb, and the motion picture camera.
  \textsc{Macaw} & 11B & Edison tv. & \xmark & No.\\
  \textsc{Macaw}-FPQ & 11B & Edison was a physicist, not a software developer. & \cmark & Tricky question. Edison was a physicist, not a computer scientist...  \\
  % , and he didn't develop software.
  \bottomrule
    \end{tabular}
}
 \caption{Additional examples of PLMs' responses to randomly selected FPQs from the test split of \OURSNOSPACE.}
\label{tab:direct_answer_examples_app}
\vspace{-0.3cm}
\end{table*}

\section{Additional Results}

\subsection{More Raw PLM's Responses to FPQs}
\label{app:dirrect_answer}
We present three more examples of PLM's responses to the FPQs and their responses to the corresponding questions that directly ask about the correctness of the premises in Table~\ref{tab:direct_answer_examples_app}. We can see that in most cases PLMs identify whether the premises are true or false successfully, however, they fail on the FPQs.

\subsection{Model-generated Answers and Explanations}
We present randomly sampled FPQs in the test split and the corresponding references, discrimination results, and explanations/answers in Table~\ref{tab:random_sample_10}. We use \textsc{Macaw-11B} trained with full training data while binary loss is added in this demonstration. We can see that in most cases, the explanation generated by the model is close to the reference. However, there are cases that the generated explanation is counterfactual. For example, ``\textit{A spider's shell is not helpful to its breath}'' is incorrect. 
\label{app:samples}

\begin{table*}[]
    \centering
\resizebox{\textwidth}{!}{
    \begin{tabular}{c|m{4.5cm}|m{5.5cm}|c|m{5.5cm}}
    \toprule
   No. &  \multicolumn{1}{>{\centering\arraybackslash}c|}{Question} & \multicolumn{1}{>{\centering\arraybackslash}c|}{Reference} & Classification & \multicolumn{1}{>{\centering\arraybackslash}c}{Explanation/Answer}\\
     \midrule
  1 &  Why are birds not a kind of animals? & By definition, birds is a type of animal.  &  \cmark & Birds are animals.   \\
  2 &Is a spider's shell helpful to its breath? & A spider has no shell. & \cmark & A spider's shell is not helpful to its breath. \\
3 & Why can't lithium burn? & Lithium burns and explodes easily. & \xmark & Because lithium is a metal.\\
4 & When do the feet of the cane grow? & Cane is not an animal, so it doesn't have feet. & \cmark & Canes are plants, not animals, and they don't have feet. \\
5 & List a type of clock that has two ears. & A clock is not an animal, it doesn't have two ears.  & \cmark & Clocks are mechanical devices, not biological, and have no ears. \\
6 & How long will it take plants to grow into adult animals? & Plants don't grow into adult animals, plants grow  into a tree or bigger plants. & \cmark & Plants are plants, animals are animals, they can't grow into plants. \\
7 & What material was the camera made of in 1350? & The camera was invented in 1839. & \cmark & The camera obscura was invented in 1839. \\
8 & Which place has eight seasons? & There are only 4 seasons in total. & \cmark & There are only four seasons. \\
9 & Why does the pregant woman have two bellies? & A pregnant woman has only one belly. & \cmark & Pregnant women have one belly. \\
10 & If foxes live in the sea, how often do they come ashore? & Foxes have no gills and cannot live in the sea. & \cmark & Foxes live on land.\\
% train_exp-3_falseqa_scale-1187_macaw-11b_2023-01-20_09-02-28_4.csv
\bottomrule
    \end{tabular}
}
    \caption{Randomly sampled FPQ questions in the test split. Column ``Reference'' presents one of the two reference explanations. Column ``Classification'' denotes whether the model successfully classifies the question as FPQ. Column ``Explanation'' gives the model-generated explanation for the rebuttal case and the model-generated answer to the non-rebuttal case.}
    \label{tab:random_sample_10}
\end{table*}

\subsection{The Questions in ARC-DA that \textsc{Macaw}-FPQ Rebuts}
We show the problem that the model still rebuts after data replay. Specifically, we show the model results for the \textsc{Macaw}-11B model after training on the full training data as well as the replayed data. Since our experiments have three seeds, we show the problem that the model refutes in all seeds. We also show the explanations generated by our model, we randomly pickle one explanation from the three seeds. As we can see in Table~\ref{tab:still_rebut}, the correctness of the  premises of these questions is not very clear. As a human, these questions can also be seen as questions containing false premises.

The question in Table~\ref{tab:still_rebut} ``\textit{How is a skin cell from a mouse similar to an amoeba?}'' can be seen as a question that contains a false premise ``\textit{A mouse's skin cells, like amoebas, are single-celled organisms.}'', as a human, we may also rebut this presupposition. For the question ``\textit{Volcanoes are considered constructive because they}'', generally, the volcanoes are considered destructive unless we want a creative answer. If a user truly wants the creative answer, he might provide explicit instructions to the PLM to trade robustness for creativity, which can be future work. 
\label{app:stillrebuts}
\begin{table*}[]
    \centering
\resizebox{\textwidth}{!}{
    \begin{tabular}{c | m{9cm}|m{10cm}}
    \toprule
  No. & \multicolumn{1}{>{\centering\arraybackslash}c|}{Question}  & \multicolumn{1}{>{\centering\arraybackslash}c}{Explanation} \\
     \midrule
1 & One explanation of why carnivores are dependent on plants is because they & Carnivores are carnivores, they don't depend on plants.\\
2 & What is responsible for changing the remains of prehistoric sea animals into natural gas? & Prehistoric sea animals were fossilized in the sedimentary rock, not in the gas form.\\
3& Which planet is fourth from the Sun in our solar system? & The fourth planet from the Sun is the Moon.\\
4 & How do sprouting plants exhibit positive gravitropism? & Plants are not animals, they are not able to gravitropise.\\
5 & Volcanoes are considered constructive because they& Volcanoes are destructive because they release lava.\\
6 & How is a skin cell from a mouse similar to an amoeba? & Amoebas are single-celled organisms, not skin cells. \\
\bottomrule
    \end{tabular}
}
    \caption{All questions from ARC-DA dataset that the model trained with the data replay technique still rebuts across all seeds. The first and fifth questions are not incomplete because they take the format of sentence continuation.}
    \label{tab:still_rebut}
\end{table*}

\end{document}